# Generalizability of CNN Architectures for Face Morph Presentation Attack

Sherko R. HmaSalah[1], Aras T. Asaad[1,2]

[1]Department of Software Engineering, Faculty of Engineering, Koya University,
Danielle Mitterrand Boulevard, Koya KOY45, Kurdistan Region – F.R. Iraq

[2]School of computing, The University of Buckingham, UK.

ABSTRACT

Automatic border control systems are wide spread in modern airports worldwide. Morphing attacks on face biometrics is a serious threat that undermines the security and reliability of face recognition systems deployed in airports and border controls. Therefore, developing a robust Machine Learning (ML) system is necessary to prevent criminals crossing borders with fake identifications especially since it has been shown that security officers cannot detect morphs better than machines. In this study, we investigate the generalization power of Convolutional Neural Network (CNN) architectures against morphing attacks. The investigation utilizes 5 distinct CNNs namely ShuffleNet, DenseNet201, VGG16, EffecientNet-B0 and InceptionResNet-v2. Each CNN architecture represents a well-known family of CNN models in terms of number of parameters, architectural design and performance across various computer vision applications. To ensure robust evaluation, we employ 4 different datasets (Utrecht, London, Defacto and KurdFace) that contain a diverse range of digital face images which cover variations in ethnicity, gender, age, lighting condition and camera setting. One of the fundamental concepts of ML system design is the ability to generalize effectively to previously unseen data, hence not only we evaluate the performance of CNN models within individual datasets but also explore their performance across combined datasets and investigating each dataset in testing phase only too. Experimental results on more than 8 thousand images (genuine and morph) from the 4 datasets show that InceptionResNet-v2 generalizes better to unseen data and outperforms the other 4 CNN models.

**Keywords**: Biometrics, CNN, Deepfake, Machine Learning, Morph Detection

# I. INTRODUCTION

Nowadays, facial images are widely used in many biometric systems to identify and authenticate individuals. With the continues advancement of Deepfake technology and image editing tools, potential threats emerge from the misuse of manipulated images and freely available fake generation platforms. Consequently, the need for reliable approaches for image analysis becomes more crucial to build a robust facial based biometric system. Morphing attacks on face biometrics is a recent threat that undermines the security of face recognition systems deployed in borders and airports worldwide. The main issue of morphed faces is the fact that many modern face recognition systems accept those fake images as genuine images (Asaad and Jassim, 2017). Furthermore, the security personnel in airports and border controls cannot easily distinguish morphed faces from their genuine counterparts (Kramer et al., 2019). This threat is even harder to detect when the created morphed faces are visually faultless and fraudsters spend a lot of time to manually remove visible artifacts. Hence, this paper addresses the issue of morphed face detection and the generalisation problem that affect many deep learning approaches and prevents them from deployment in practise.

Typically, a morphed face is a face that contain face features of two or more persons and looks like two or more persons. In general, morphing is a process of combining two or more faces by first aligning the frontal faces, and then averaging the corresponding intensity pixel values. But different methods have been proposed over the years to make the final generated morph image to be more realistic. In this vein, combined morph, complete morph, splicing morph, OpenCV are among four well-known landmark-based approaches of creating morphed faces. The first three methods have been proposed by (Makrushin et al., 2017), (Neubert et al., 2018) and OpenCV implementation can be seen in (Mallick, 2016) .Deep learning algorithms, on the other hand, can also be used to generate high quality morphs known as Generative Adversarial Networks (GAN) and especially StyleGAN (Sarkar et al., 2022). Beside these, there are approaches whereby the underline morphing algorithm is not known such as web-based morphing generation tools like (FaceMorpher), (MorphThings), (FaceShape) and others.

Another issue in detecting morphed faces is known as print-scan issue because in many countries the passport applicant needs to provide a printed copy of their face and the passport issuance officer will then scan it to save a copy in the database and printed on the passport too. This process of printing and then scanning the face distorts tiny local features which makes the classification task more difficult by algorithms as well as humans. Hence, there is an immediate need to design robust and sophisticated algorithms that can detect morphed faces and help security officers to prevent unauthorized people to cross borders. This work is an attempt to investigate the effectiveness of 5 well known CNN architectures to detect morphed faces, as well 4 different datasets whereby each dataset morphing process is unique. The main contributions of this paper can be summarized as follows:

- What is the effect of unknown morphing generation methods on the effectiveness of Morphing Attack Detections (MAD) systems? To address this issue, we utilized 4 different morph generation approaches known as Combined (Utrecht Dataset), Splicing (Defacto Database), OpenCV (KurdFace dataset) and StyleGAN (AMSL database).
- Testing the Generalization power of 5 well-known CNN architectures (ShuffleNet, DenseNet, EfficientNet-B0, VGG16 and InceptionResNet-V2) by imposing two conditions. First, training and testing datasets are different (cross-dataset) and secondly the morphing generation algorithm in training is different from the testing. This is mainly to mimic the real application of MAD in border control scenarios.
- Investigating the effectiveness of 5 popular CNN architectures to detect morphing attacks. We opt to use ShuffleNet, DenseNet-201, EfficientNet-B0, VGG16 and InceptionResNet-V2 such that each model represents a different family of CNN architectures that works best in many computer vision applications, especially ImageNet, CIFAR100 and CIFAR10 natural database benchmarks.

The rest of this paper organized as follows. section II contains a literature review of related work. The CNN architectures briefly described in section III. Utilized databases and their description reported in section IV. Experimental design and testing protocols are presented in section V. The followed section highlights conclusion and future research directions.

## II. LITURATURE REVIEW

Many efforts have been paid from the researcher community to tackle the vulnerability of face recognition systems to morph attacks. In 2014 (Ferrara et al., 2014), first reported the threat of morphing attacks on face biometrics. Then a large number of computer algorithms have been proposed for MAD and creating morphing algorithms.

In 2021 (Banerjee and Ross, 2021) has proposed an unsupervised learning model by using a conditional generative network (cGAN) to detect morphed faces conditioned on the reference image, in intra and cross datasets. Their experiments were conducted on London, MorGAN, and E-MorGAN datasets. The best outcome reported was 3% Bona Fide Presentation Classification Error Rate (BPCER) @ 10% Attack Presentation Classification Error Rate (APCER) on intra-dataset and 4.6% BPCER @ 10% APCER on cross dataset.

In another study (Chaudhary et al., 2021) utilized 2D Discrete Wavelet Transform (DWT) by calculating the entropy of multi-level sub-bands then using the Kullback-Liebler Divergence (KLD) measurement to discriminate 22 sub-bands out of 48 produced sub-bands, the isolated sub-bands are fed to a deep Siamese Neural Network. They conducted various experiments with different results. The performance of their proposed method using APCER @ 5% BPCER is 33.78 and D-EER is 16.4 when AMSL dataset used in testing only while the training was other datasets. (Kraetzer et al., 2021) have investigated the effect of fusion to detect morphing faces in passport scenarios. They used 5 algorithms, 1 landmark-based approach and 4 VGG19 CNN variations, and concluded that a naïve approach of fusion will not boost the overall MAD performance. They tested different fusion scenarios such as majority voting, weighted linear combination, Fusion based on Dempster-Shafer Theory and fusion using likelihood ratios.

Furthermore,(Hussein and Rashid, 2022) have created their own face database, called KurdFace dataset, as well as generated their corresponding OpenCV based morphs. They tested the performance of Local Binary Patterns (LBP) and Uniform LBP (ULBP) on KurdFace and AMSL dataset. They used different training and testing protocols such as 50% training/testing, 75% training and 25% testing in a balanced setting. For both datasets, the reported performance for BPCER and APCER is (on average) over 10% and 15% respectively for both datasets using either LBP or ULBP. The concept of persistent homology adapted by (Jassim and Asaad, 2018) for the purpose of MAD where they used ULBP to select 0-dimentional simplices and then creating a filtration to obtain what is known as persistence barcodes. They used different databases in their investigation and achieve good performance when using the same dataset in training and testing but they have not reported any generalization performance (i.e. using an unseen database in testing). For a systematic review and survey of recent morphing attack detection and morphing creation techniques, we direct interested readers to see (Kenneth et al., 2022) and (Hamza et al., 2022).

Aforementioned literature lacks addressing the generalization problem of CNN architectures for the detection of morphed face images using a diverse family of CNN models. Especially, using CNN models with small and large number of trainable parameters and diversity in the CNN architecture design.

### III. CNN ARCHITECTURES

CNN is a pioneering breakthrough recorded a revelational milestone in various fields such as pattern recognition, computer vision, artificial intelligence, by leveraging its capability to derive automatic features from intricate data, paved the path for significant advancements in image classification, image segmentation, object detection, medical imaging analysis, and many more.

To the best of our knowledge, the history of CNN goes back to (Fukushima, 1980) where he proposed a neural network architecture named Neocognitron as a first model in the field of visual pattern recognition. Neocognitron organizes itself and capable of identifying stimulus by considering their geometrical similarities. Nine years later, (LeCun et al., 1989) has announced the development of a fresh architectural design which eventually gained popularity and recognized as LeNet. The architecture employs back-propagation (BP) to recognize hand-written digits. Subsequently LeNet gained widespread popularity and the majority modern CNN architectures inspired by the advancements and foundations of LeNet and back propagation. Hence, convolutional layers and BP became essential components of modern deep learning design. In 2012, a deep learning model, known as AlexNet (Krizhevsky et al., 2012), won the ImageNet competition and its where the CNN gained more popularity. It was the first CNN which has a deep network that constitutes of many convolutional layers and the authors deployed the power of Graphical Processing Unit (GPU) in their training protocols.

(Simonyan and Zisserman, 2014) in oxford university, introduced VGG16 to investigate how the accuracy of large-scale image being affected by the depth of a CNN, the investigation used 16 layers including 13 convolutional layers and 3 Fully-Connected (FC) layers (VGG19 having 16 convolutional and 3 FC layers). The convolutional layers used a small 3x3 convolutional filters and a 2x2 window size with stride 2 for max pooling layers. ResNet (He et al., 2016) family of CNN architecture, on the other hand, use the concept of residual connections to address the issue of weight vanishing and ResNet18 and ResNet50 are among the popular adapted architectures.

Convolutional filters in traditional CNNs used filters in large size to capture spatial information, this may result in memory consumption and increasing computational time, to address these problems, InceptionNet was developed by (Szegedy et al., 2015) at Google as a part of the GoogLeNet project. GoogLeNet utilized 9 inception modules with 22 layers deep as its architecture, and in 2014 it achieved remarkable accomplishment in the ImageNet Large-Scale Visual Recognition Challenge (ILSVRC) competition. Later, InceptionResNet-v4 (Szegedy et al., 2017) has been introduced where they combine the inception and residual connection concepts within the same architecture. DenseNet-201 mode is a variant of DenseNet which was proposed by (Huang et al., 2017), the model consist of 201 fully connected layers i.e. each layer receives the output of all preceding layers and passes its own output to all subsequent layers. This connectivity offers vanishing-gradient alleviation, reuse features encouragement, and strengthen feature propagation. The model evaluation on (CIFAR-10, CIFAR-100, SVHN, ImageNet) benchmark datasets showed significant improvement compared the state-of -the-art at the time.

To address the problem of accuracy and computational efficiency of deep neural networks on resource-limited devices such as mobile and embedded devices, (Zhang et al., 2018) in 2017 developed a novel architecture named ShuffleNet. The fundamental principle of this architecture involves employing pointwise group convolution and channel shuffling. They reported that this architecture outperforms MoblieNet on ImageNet classification task and scored higher speed compared to AlexNet while achieving similar accuracy. (Tan and Le, 2019) from Google Research developed EfficientNet2 as an improved version of the original of EfficientNet architecture, the researchers proposed a novel approach by introducing a new scaling parameter called compound coefficient that uniformly scales all dimensions including depth, width and resolution without accuracy and efficiency being affected. Furthermore, they reported that their proposed method is about 8.4x smaller and 6.1x faster than the best existing CNN architecture.

In this work, we employ VGG16, DenseNet-201, ShuffleNet, EfficientNet-B0 and InceptionResNet-v2 whereby each model represents a family of the CNN architectures exist in literature ranging from models with large number of trainable parameters (VGG16, InceptionResNet-v2) to small number of trainable parameters (ShuffleNet, EffecientNet-B0, DenseNet-201). All CNN models have been used in a transfer learning mode where we fine tune all hidden layers as well as convolutional layer weights. We used pretrained CNN models in MATLAB Version R2022b with the following hyperparameter settings using **trainingOptions** function: Adam optimiser, 10 epochs, 12 minibatch Size, 1e-4 InitialLearnRate and a ValidationFrequency of 30. We also used **imageDataAugmenter** function in MATLAB for augmentation by setting up parameters to: [-5,5] for RandRotation, RandXReflection and RandYReflection with value of 1, and using [-0.05 0.05] for both RandXShear, RandYShear.

IV. DATASETS, MORPHING TECHNIQUES AND TESTING PROTOCOLS

To evaluate the performance of the CNN models we proposed to utilize in this work, we used 4 different datasets namely KurdFace, London, Utrecht and Defacto. These datasets contain genuine as well as their corresponding generated morphed faces. Firstly, KurdFace dataset created by Salahaddin University researchers, in Kurdistan region of Iraq, where they collected 170 genuine faces mostly from university students and lecturers. They then generated corresponding morphed faces using OpenCV library in python and finally they kindly shared with us 2000 manually selected good looking morphs. London dataset contain 102 genuine faces from the Face Research Lab London Set(FRLS) (DeBruine and Jones, 2017) and then 1222 high quality morphs are generated using StyleGAN approach (Sarkar et al., 2022). The morphs generated using StyleGAN are based on the new advancements in the field of GAN and in particular styleGAN2 (Karras et al., 2020).

In general, GAN is a type of deep learning model that constitutes of two networks competing against each other known as generator and discriminator. The generator network learns to generate fake data, while the discriminator network learns to distinguish the generator's fake data from real data. The discriminator network penalizes the generator for producing unrealistic results. When training begins, the generator produces obviously fake data (mostly noise) and the discriminator then quickly learns to recognize that it's fake. The generator network then adjusts its learnable weights to produce better data and the discriminator adjusts its parameters to better distinguish between real and fake data. This process will continue until the generator produces data that is indistinguishable from real data. StyleGAN, on the hand, is a type of GAN such that its generator network architecture includes a mapping network that maps points in latent space to an intermediate latent space, which is used to control the style of the generated images. As a source of variation, the generator network also introduces noise at each point in the generator model. The generator network no longer takes a point from the latent space as input. Instead, there are two new inputs: a style vector and a noise vector. The style vector is used to control the style of the generated image, while the noise vector is used to add stochastic variation to the image. Full details of how GAN and StyleGAN work can be found in (Karras et al., 2019) (Goodfellow et al., 2014) . StyleGAN based morphs for the 102 London faces are obtained from (Sarkar et al., 2022).

Utrecht dataset created from 67 genuine faces from (ECVP) and 2651 morphed faces generated using combined morphing technique in (Neubert et al., 2018) .In a nutshell, combined morphing technique warps the facial regions first then averages the intensity pixel values and then stitching it back on the warped image. To address artefacts, especially between the forehead and frontal face region, poisson image editing will be applied to the entire image to produce a visually faultless morph. Morphs generated using combined morph has an average texture and geometry from both contributing faces and skin color has no influence. Finally, Defacto (Mahfoudi et al., 2019) dataset contain 200 genuine faces and more than 39000 morphed faces generated using splicing technique. Splicing is a landmark-based approach that uses 77 face landmarks to cut a convex hull from both input faces to the morphing pipeline. It then uses Delaunay triangulation, using average landmark coordinates, to warp the two faces into an average position and then computes average intensity pixel values. Generated morphed frontal face is then warped back to original images. The limitation of splicing morph is the fact that it relies on the geometry of only one of the input faces and produces minor ghosting artifacts. We randomly selected 2309 images to reduce the highly imbalance bias between genuine and morphing samples in Defacto dataset. Table I contain statistics of genuine and morphed images from the four datasets utilized in this paper. Figure 1 shows a selection sample of bona fide with their morphed images of each dataset.

*Table I*
*DATASET DESCRIPTION FOR THIS STUDY*

| DATASET | BONA FIDE | MORPH | MORPH TYPE |
|---|---|---|---|
| **KURDFACE** | 170 | 2000 | OpenCV |
| **LONDON (AMSL)** | 102 | 1222 | StyleGAN |
| **UTRECHT** | 67 | 2651 | Combine |
| **DEFACTO** | 200 | 2309 | Splicing |

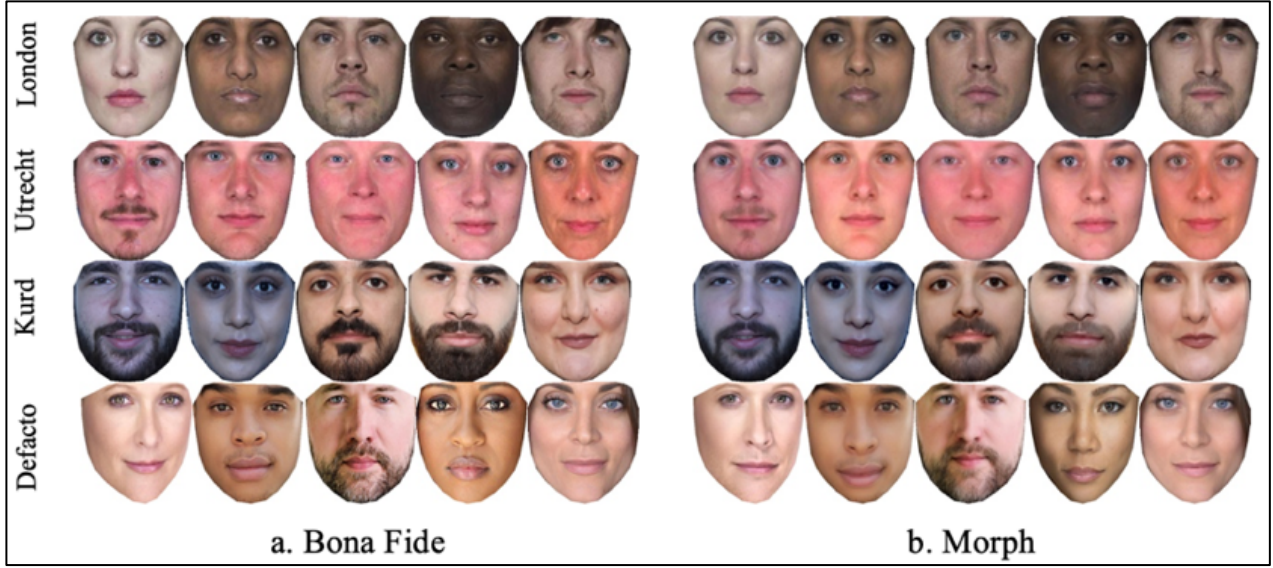

*Figure I: Samples of Dataset images (a) Bona Fide faces and the corresponding (b)Morphed faces.*

To evaluate the capabilities of CNNs, three different evaluation scenarios have been applied on the datasets. The **first scenario** is conducting tests on each dataset separately by using 5-fold cross validation, then followed by **second scenario** which is mixing the four datasets and then performing 5-fold cross validation. Finally, the **third scenario** is to test the generalizability power of CNNs by using 3 datasets in training and using one dataset in testing. When a dataset is used in testing, its corresponding genuine and morphed images will be in testing. This is to address two issues in MAD systems. First, to see if the morphing technique in testing is different from the training morphing techniques, how the CNN models perform? This is to address the generalizability for different (unseen) morphing techniques. Secondly, using different face datasets in training and testing provides more guarantee that the model performance may generalize to other unseen datasets.

## V.  EVALUATION METRICS AND EXPERIMENTAL RESULTS

This section presents results and main findings of this study in order to evaluate the performance of the 5 CNN architectures. We use three performance measurements known as APCER, BPCER, and Detection Equal Trade-off (DET). APCER refers to the proportion of attack presentations incorrectly classified as bona fide presentations (see equation 1) while BPCER is the proportion of bona fide presentations which are misclassified as attack presentations (see equation 2).

$$APCER = FN/(TP + FN) \quad (1)$$
$$BPCER = FP/(FP + TN) \quad (2)$$

Where FN, TP, FP and TN are the total number of false negatives, total number of true positives, total number of false positives and total number of true negatives, respectively. A false negative means that a morph face incorrectly classified as bona fide face and false positive refers to a bona fide face misclassified as morphed image. True positive means a morphed image correctly recognized as morph whereas true negative is the number bona fide faces correctly classified as bona fide.

Then we used APCER and BPCER to calculate Detection Error Trade-off (DET) which is another measurement that can be used to evaluate the efficiency of each CNN model, i.e. the DET is a point in which the rate of APCER is equal to BPCER. Finally, we also report APCER at 5% BPCER which measures APCER at 5% while maintaining a low false rejection rate (i.e. BPCER = 5%).

Next, we are demonstrating the results we obtained following each of the training scenarios proposed in the previous section. In table II and table III, we show the results we obtained using the images in the same dataset for training and testing in 5-fold cross validation setting. We use the images in the 4 datasets as it is in the training and testing, i.e. with the highly unbalanced number of morphs compared to bona fide images.

TABLE II:
INTRA DATASET PERFORMANCE OF KURDFACE AND LONDON DATASET FOR THE 5 CNN MODELS.

|  | KurdFace (Train/Test) | | | | London (Train/Test) | | | |
| --- | --- | --- | --- | --- | --- | --- | --- | --- |
| CNN ARCHITECTURES | APCER | BPCER | APCER @ 5% BPCER | DET | APCER | BPCER | APCER @ 5% BPCER | DET |
| SHUFFLENET | 0.15 | 0 | 99.25 | 100 | 0.16 | 0 | 98.61 | 100 |
| INCEPTIONRESNETV2 | 0.05 | 1.76 | 99.80 | 99.1 | 0 | 0 | 99.92 | 100 |
| EFFICIENTNET-B0 | 0 | 0 | 99.95 | 100 | 0 | 0 | 99.10 | 100 |
| VGG16 | 0 | 100 | 91.25 | 54.4 | 1.8 | 42 | 52.37 | 79.6 |
| DENSENET201 | 0.15 | 0 | 98.85 | 100 | 0 | 0 | 99.84 | 100 |

Table III:
INTRA DATASET PERFORMANCE OF UTRECHT AND DEFACTO DATASET FOR THE 5 CNN MODELS.

|  | Utrecht (Train/Test) | | | | Defacto (Train/Test) | | | |
| --- | --- | --- | --- | --- | --- | --- | --- | --- |
| CNN ARCHITECTURES | APCER | BPCER | APCER @ 5% BPCER | DET | APCER | BPCER | APCER @ 5% BPCER | DET |
| SHUFFLENET | 0.08 | 5.97 | 99.06 | 97.3 | 0.82 | 5.00 | 96.32 | 96.54 |
| INCEPTIONRESNETV2 | 0 | 13.43 | 100 | 99.22 | 1.34 | 1 | 94.80 | 99 |
| EFFICIENTNET-B0 | 0.23 | 4.48 | 99.47 | 97.85 | 0.95 | 3.5 | 96.71 | 98.03 |
| VGG16 | 0 | 100 | 100 | 49.41 | 8.23 | 21 | 72.85 | 80 |
| DENSENET201 | 0 | 1.49 | 100 | 100 | 0.95 | 1 | 94.98 | 99 |

It is clear from table II that all CNN models are performing well on KurdFace and London dataset,

except VGG16 which performs poorly on both datasets. We suspect that VGG16 architecture is too complex for this task of morph detection and the number of learnable parameters is 138 million which is more than double of InceptionResNet-v2 and far more than the other 3 CNN models which are light weight architectures. Using the same intra dataset training and testing approach, it can be noticed from table III that DenseNet-201outperforms all other CNN models with 0% APCER and 1.49% BPCER in Utrecht dataset, followed by EffecientNet-B0 and ShuffleNet. Note that it is the light weight CNN models which perform well and the two CNN architecture with high number of parameters, VGG16 and InceptionResNet-v2, perform the worst. In the same vein, DenseNet-201 outperformed the rest of CNN models with 0.95% APCER and 1% BPCER in Defacto dataset followed by InceptionResNet-v2 and EffecientNet-B0. Again, VGG16 performed the worst in Defacto database.

Following the second scenario in the evaluation protocol, we mixed the four datasets, and then applied 5-fold cross validation to train and test the 5 CNN architectures of interest. In table IV, we report the results we obtained using the 4 evaluation metrics defined in the previous section. It is clear that EffecientNet-B0 outperforms the other CNN models with 0.8% APCER and 1.48% BPCER. Then followed by DenseNet201 (0.9% APCER, 2.23% BPCER) and ShuffleNet (2.4% APCER, 3.71% BPCER). Note that it is the CNN architectures with small number of parameters (i.e. weights) which are performing well when mixing the four datasets.

*Table IV:*
*MIXING ALL DATASET PERFORMANCE FOR THE 5 CNN MODELS.*

| CNN ARCHITECTURES | APCER | BPCER | APCER @ 5% BPCER | DET |
|---|---|---|---|---|
| SHUFFLENET | 2.4 | 3.71 | 88.6 | 96.87 |
| INCEPTIONRESNETV2 | 0.5 | 10.2 | 98.6 | 97.64 |
| EFFICIENTNET-B0 | **0.8** | **1.48** | 94.9 | 98.89 |
| VGG16 | 4.1 | 41.19 | 49.5 | 77.23 |
| DENSENET201 | 0.9 | 2.23 | 95.9 | 98.7 |

Finally, following the third scenario in the evaluation protocol, we use train each CNN model using 3 datasets and use one dataset in testing, in turn so that each dataset will be tested consequently. In table V and VI, we show the results for when each of the datasets are in testing phase.

*Table V:*
*CROSS DATASET RESULTS. USING KURDFACE AND LONDON DATASETS IN TESTING ONLY.*

| | KurdFace (Test Only) | | | | London (Test Only) | | | |
|---|---|---|---|---|---|---|---|---|
| CNN ARCHITECTURES | APCER | BPCER | APCER @ 5% BPCER | DET | APCER | BPCER | APCER @ 5% BPCER | DET |
| SHUFFLENET | 0 | 1.76 | 99.55 | 99.43 | 4.01 | 24.51 | 86.5 | 88.19 |
| INCEPTIONRESNETV2 | **1.35** | **0** | 94.9 | 98.81 | **6.38** | **3.92** | 82 | 95.05 |
| EFFICIENTNET-B0 | 3.7 | 1.18 | 81.04 | 98.71 | 0 | 37.26 | 99.67 | 94.16 |
| VGG16 | 0.1 | 58.82 | 97.05 | 89.25 | 59.33 | 4.90 | 20.79 | 80.09 |
| DENSENET201 | 2.1 | 2.35 | 95.2 | 97.75 | 0.08 | 50 | 99.1 | 95.1 |

In table V, it can be noticed that for when either KurdFace or London database are in testing, InceptionResNet-v2 outperforms the other 4 CNN models with 1.35% APCER, 0% BPCER on KurdFace dataset and 6.38% APCER and 3.92% BPCER on London database.

Finally, in table VI we show the results we obtained using Utrecht and Defacto database in testing phase. EffecientNet-B0 outperforms the other 4 CNN models in Utrecht database while InceptionResNet-v2 outperforms the other CNN architectures in Defacto database. If one wants to select an architecture that performs well regardless of which database is in testing phase, then the winner will be InceptionResNet-v2 architecture.

*Table VI:*
*CROSS DATASET RESULTS. USING UTRECHT AND DEFACTO DATASETS IN TESTING ONLY*

| CNN ARCHITECTURES | Utrecht (Test Only) | | | | Defacto (Test Only) | | | |
| --- | --- | --- | --- | --- | --- | --- | --- | --- |
| | APCER | BPCER | APCER @ 5% BPCER | DET | APCER | BPCER | APCER @ 5% BPCER | DET |
| **SHUFFLENET** | 0 | 64.17 | 99.70 | 79.43 | 51.58 | 2 | 29.54 | 82.21 |
| **INCEPTIONRESNETV2** | 0.75 | 28.36 | 96.70 | 91.64 | **17.71** | **8.5** | 67.65 | 88.3 |
| **EFFICIENTNET-B0** | **0.33** | **20.89** | 96.15 | 92.98 | 30.40 | 4.5 | 45.39 | 90.76 |
| **VGG16** | 0.03 | 88.06 | 99.69 | 73.22 | 55.26 | 1 | 28.58 | 87.7 |
| **DENSENET201** | 0 | 74.62 | 99.92 | 83.39 | 36.29 | 1 | 42.92 | 89.3 |

Results obtained in this work cannot be directly compared to others in the literature because previous work, to the best of our knowledge, have not used these 4 datasets together with the exact training and testing protocols. However, the closest work we can compare our results with is that of (Hussein, 2023) and (Chaudhary et al., 2021) . In (Hussein, 2023) table 4.29 (page 94), they trained 4 CNN architectures ( AlexNet, GoogleNet, ResNet50 and SqueezeNet) using London database with combined morph generation approach and used KurdFace dataset in the testing phase. The best result they obtained is 7.1% BPCER and 5.06% APCER. Our InceptionResNet-v2 model outperforms (Hussein and Rashid, 2022) by more than 5% in both APCER and BPCER. Using LBP with different blocking scenarios, i.e. blocking the image and then computing LPB, they obtained 34.43% APCER and 8.16 BPCER when the image divided into 4 blocks. Again, our results in table V outperforms them by more than 30% in APCER and more than 8% in BPCER. Finally, they also used London (with combined morph) in the testing phase when KurdFace used in training only and they obtained 18.82% BPCER and 3.92% APCER when the image divided into 9 blocks. Similarly, our results using InceptionResNet-v2 outperform their method by more than 2% in APCER and more than 18% in BPCER.

Comparing to (Chaudhary et al., 2021) , as discussed in literature review section, they obtained 33.78% APCER @ 5% BPCER and 16.4 % D-EER when London (i.e. AMSL) dataset used in testing only while the training was Utrecht, MORGAN and LMA datasets. Despite the fact that their morphs generated using combined morphing approach, which we believe is easier to detect, our results using London (Style GAN) is 82% APCER @ 5% BPCER with 95% D-EER which outperforms their results noticeably.

Note that in our experiments, we have two extra datasets in the training phase with an extra 2 different morphing algorithms which helped our CNN models to learn more and generalize better when either KurdFace or London (referred to as AMSL in (Hussein, 2023)) used in the testing phase. Hence, this is an indication that a diverse set of datasets with different morphing approaches helps in boosting the MAD performance.

## VI. CONCLUSION AND FUTURE RESEARCH DIRECTIONS

In this paper, we investigated the use of 5 different CNN models for the purpose of single MAD task using 4 different datasets and 4 different morph generation algorithms. Each selected CNN architecture represents a family of the CNN models developed for computer vision tasks. In particular, we selected VGG16 to represent VGG (and AlexNet) family of CNN models, InceptionResNet-v2 to represent ResNet and Inception/Google family of CNN models, EffecientNet to represent the EffecientNet family, ShuffleNet to represent light-weight architectures (1.2 million parameters) and finally DenseNet-201 to represent DenseNet models. All utilized CNN architectures have been used in transfer learning mode with fine tuning all parameters in MATLAB. To test the effectiveness of selected CNN algorithms, we designed 3 evaluation protocols from easy to difficult settings that mimics real life scenarios adopted in border controls and airports. Especially third evaluation protocol whereby we used an unseen dataset in testing phase such that its morphed images generated using a different technique than the methods used to generate the morphs in the training. This test sheds light into the generalizability power of each CNN architecture as well as the effect of using different morphing approach in training and testing.

Furthermore, we opt to use landmark-based morph generation methods as well as non-landmark-based methods such as StyleGAN to address the unknown morph generation algorithm issue in MAD. The 4 datasets we used in our investigation reflects a diverse range of ethnicities as well as diverse lighting conditions, closeness to the camera, gender and age balance.

In general, using the same dataset or mixing more than one dataset and then splitting training and testing is somehow easy for CNN models to distinguish genuine faces from their morph counterparts, see table II, III and IV. On the other hand, using a different (unseen or cross) dataset in testing phase with different morph generation is the most difficult scenario one needs to consider to prepare an applicable MAD system to be deployed in airports. In this vein, the results in table V and VI suggests that an architecture, such as InceptionResNet-v2, with a medium number of parameters (55 million parameters) and architecture complexity maybe a good candidate to detect MAD that generalizes well on unseen data. In other words, a very light weight CNN architecture like ShuffleNet with 1.4 million parameter or VGG16 architecture with 138 million parameters may not be a good model to generalize well on unseen data. It is clear that combining inception and residual concepts in one architecture is a good starting point to design a sophisticated single MAD system.

There are a number of limitations in the current work. One can further optimize InceptionResNet-v2 architecture by optimizing hyperparameter tunning or redesigning it by taking inspirations from the architecture itself. Secondly, we have not tested other morphing generation algorithms such as complete morph, FaceMorpher and others. We also have not tested the effect having a balanced number of images in training and testing which we believe is an easier setting than imbalance scenario. Finally, we have not considered the explainability of utilized CNN models to check which part of the image has been used the most by CNNs.

In addition to addressing each of the aforementioned limitations, current work also opens the door for further research work. How does each of the CNN models utilized in this work perform on print-scanned scenario? Especially as we know that print-scan process distorts local features and also different scanning devices can have an effect on the MAD. Assembling a dataset which is more diverse especially to contain more African, Latin and Asian ethnicity individuals which might not be an easy task but would be necessary for a good MAD system design.

# REFERENCES

Asaad, A. & Jassim, S. Topological data analysis for image tampering detection. Digital Forensics and Watermarking: 16th International Workshop, Magdeburg, Germany, Proceedings 16, 2017. Springer, 136-146.

Banerjee, S. & Ross, A. Conditional identity disentanglement for differential face morph detection. 2021 IEEE International Joint Conference on Biometrics (IJCB), 2021. IEEE, 1-8.

Chaudhary, B., Aghdaie, P., Soleymani, S., Dawson, J. & Nasrabadi, N. M. Differential morph face detection using discriminative wavelet sub-bands. Proceedings of the IEEE/CVF Conference on Computer Vision and Pattern Recognition, 2021. 1425-1434.

Debruine, L. & Jones, B. 2017. Face research lab London set. *Psychol. Methodol. Des. Anal*.

Ecvp, U. Available: https://pics.stir.ac.uk/2D_face_sets.htm [Accessed July 26 2023].

Facemorpher. Available: https://andy6804tw.github.io/FaceMorpher/ [Accessed July 26 2023].

Faceshape. Available: https://www.faceshape.com/face-morph [Accessed July 26 2023].

Ferrara, M., Franco, A. & Maltoni, D. 2014. The magic passport. *IEEE International Joint Conference on Biometrics.* Clearwater, FL,USA.

Fukushima, K. 1980. Neocognitron: A self-organizing neural network model for a mechanism of pattern recognition unaffected by shift in position. *Biological cybernetics,* 36**,** 193-202.

Goodfellow, I., Pouget-Abadie, J., Mirza, M., Xu, B., Warde-Farley, D., Ozair, S., Courville, A. & Bengio, Y. 2014. Generative adversarial nets. *Advances in neural information processing systems,* 27.

Hamza, M., Tehsin, S., Humayun, M., Almufareh, M. F. & Alfayad, M. 2022. A comprehensive review of face morph generation and detection of fraudulent identities. *Applied Sciences,* 12**,** 12545.

He, K., Zhang, X., Ren, S. & Sun, J. Deep residual learning for image recognition. Proceedings of the IEEE conference on computer vision and pattern recognition, 2016. 770-778.

Huang, G., Liu, Z., Van Der Maaten, L. & Weinberger, K. Q. Densely connected convolutional networks. Proceedings of the IEEE conference on computer vision and pattern recognition, 2017. 4700-4708.

Hussein, A. R. 2023. *Single Morph Attack Detection Using Machine Learning.* MSc., Salahaddin.

Hussein, A. R. & Rashid, R. D. 2022. KurdFace morph dataset creation using OpenCV. *Science Journal of University of Zakho,* 10**,** 258-267.

Jassim, S. & Asaad, A. Automatic detection of image morphing by topology-based analysis. 26th European Signal Processing Conference (EUSIPCO), 2018. IEEE, 1007-1011.

Karras, T., Laine, S. & Aila, T. A style-based generator architecture for generative adversarial networks. Proceedings of the IEEE/CVF conference on computer vision and pattern recognition, 2019. 4401-4410.

Karras, T., Laine, S., Aittala, M., Hellsten, J., Lehtinen, J. & Aila, T. Analyzing and improving the image quality of stylegan. Proceedings of the IEEE/CVF conference on computer vision and pattern recognition, 2020. 8110-8119.

Kenneth, M. O., Sulaimon, B. A., Abdulhamid, S. M. & Ochei, L. C. 2022. A systematic literature review on face morphing attack detection (mad). *Illumination of Artificial Intelligence in Cybersecurity and Forensics***,** 139-172.

Kraetzer, C., Makrushin, A., Dittmann, J. & Hildebrandt, M. 2021. Potential advantages and limitations of using information fusion in media forensics—a discussion on the example of detecting face morphing attacks. *EURASIP Journal on Information Security***,** 1-25.

Kramer, R. S., Mireku, M. O., Flack, T. R. & Ritchie, K. L. 2019. Face morphing attacks: Investigating detection with humans and computers. *Cognitive research: principles and implications,* 4**,** 1-15.

Krizhevsky, A., Sutskever, I. & Hinton, G. E. 2012. Imagenet classification with deep convolutional neural networks. *Advances in neural information processing systems,* 25.

Lecun, Y., Boser, B., Denker, J., Henderson, D., Howard, R., Hubbard, W. & Jackel, L. 1989. Handwritten digit recognition with a back-propagation network. *Advances in neural information processing systems,* 2.

Mahfoudi, G., Tajini, B., Retraint, F., Morain-Nicolier, F., Dugelay, J. L. & Marc, P. DEFACTO: Image and face manipulation dataset. 27Th european signal processing conference (EUSIPCO), 2019. IEEE, 1-5.

Makrushin, A., Neubert, T. & Dittmann, J. Automatic generation and detection of visually faultless facial morphs. International conference on computer vision theory and applications, 2017. SciTePress, 39-50.

Mallick, S. MARCH 11, 2016. *Face Morph Using OpenCV — C++ / Python* [Online]. Available: https://learnopencv.com/face-morph-using-opencv-cpp-python/ [Accessed JULY 26 2023].

Morphthings. Available: https://www.morphthing.com [Accessed July 26 2023].

Neubert, T., Makrushin, A., Hildebrandt, M., Kraetzer, C. & Dittmann, J. 2018. Extended StirTrace benchmarking of biometric and forensic qualities of morphed face images. *IET Biometrics,* 7**,** 325-332.

Sarkar, E., Korshunov, P., Colbois, L. & Marcel, S. Are GAN-based morphs threatening face recognition? IEEE International Conference on Acoustics, Speech and Signal Processing (ICASSP), 2022. IEEE, 2959-2963.

Simonyan, K. & Zisserman, A. 2014. Very deep convolutional networks for large-scale image recognition. *arXiv preprint arXiv:1409.1556.*

Szegedy, C., Ioffe, S., Vanhoucke, V. & Alemi, A. Inception-v4, inception-resnet and the impact of residual connections on learning. Proceedings of the AAAI conference on artificial intelligence, 2017.

Szegedy, C., Liu, W., Jia, Y., Sermanet, P., Reed, S., Anguelov, D., Erhan, D., Vanhoucke, V. & Rabinovich, A. Going deeper with convolutions. Proceedings of the IEEE conference on computer vision and pattern recognition, 2015. 1-9.

Tan, M. & Le, Q. Efficientnet: Rethinking model scaling for convolutional neural networks. International conference on machine learning, 2019. PMLR, 6105-6114.

Zhang, X., Zhou, X., Lin, M. & Sun, J. Shufflenet: An extremely efficient convolutional neural network for mobile devices. Proceedings of the IEEE conference on computer vision and pattern recognition, 2018. 6848-6856.